# Enhancing QA Systems with Complex Temporal Question Processing Capabilities

**Estela Saquete** STELA@DLSI.UA.ES
**Jose L. Vicedo** VICEDO@DLSI.UA.ES
**Patricio Martínez-Barco** PATRICIO@DLSI.UA.ES
**Rafael Muñoz** RAFAEL@DLSI.UA.ES
**Hector Llorens** HLLORENS@DLSI.UA.ES
*Natural Language Processing and Information System Group*
*Department of Software and Computing Systems*
*University of Alicante*
*Apartado de Correos 99, E-03080 Alicante, Spain*

## Abstract

This paper presents a multilayered architecture that enhances the capabilities of current QA systems and allows different types of complex questions or queries to be processed. The answers to these questions need to be gathered from factual information scattered throughout different documents. Specifically, we designed a specialized layer to process the different types of temporal questions. Complex temporal questions are first decomposed into simple questions, according to the temporal relations expressed in the original question. In the same way, the answers to the resulting simple questions are recomposed, fulfilling the temporal restrictions of the original complex question. A novel aspect of this approach resides in the decomposition which uses a minimal quantity of resources, with the final aim of obtaining a portable platform that is easily extensible to other languages. In this paper we also present a methodology for evaluation of the decomposition of the questions as well as the ability of the implemented temporal layer to perform at a multilingual level. The temporal layer was first performed for English, then evaluated and compared with: a) a general purpose QA system (F-measure 65.47% for QA plus English temporal layer vs. 38.01% for the general QA system), and b) a well-known QA system. Much better results were obtained for temporal questions with the multilayered system. This system was therefore extended to Spanish and very good results were again obtained in the evaluation (F-measure 40.36% for QA plus Spanish temporal layer vs. 22.94% for the general QA system).

## 1. Introduction

Nowadays, it is a fact that there is a huge amount of digital information available (mainly in textual form) and also a large number of users who want the easiest possible access to this information. This situation continuously fosters research on the development of information systems that make it possible to analyze, locate, manage, access and process all this information automatically. Commonly, these systems are referred to as "search engines".

A search engine is especially useful to obtain a specific piece of information without the need to manually go through all the available documentation related to the search topic. Search engines are currently evolving towards a new generation of engines capable of understanding user needs better ("the necessity behind every query") and offering specific services or interfaces, depending on the domain or context. The new generation of search engines will be able to not only offer a list of ordered web pages, but also discover pieces of information scattered throughout different information sources or even summaries (Barzilay, Elhadad, & McKeown, 2002). That is, they will integrate information from text search (web pages, documents), multimedia search (images, video, audio) and database search (tourist, biomedicine, etc.) into comprehensible answers to be delivered to users. In addition, they will correctly process questions formulated in free natural language as opposed to keyword queries or fixed templates, as in information extraction scenarios (Michelson & Knoblock, 2008). Question answering systems (QA) are one of the best examples of this new generation of search engines, allowing users to formulate questions in free natural language (NL) and providing them with exactly the information required, also in NL form.

However, QA is not a mature technology and current systems are mainly focused on the treatment of questions that require very specific items of data as an answer such as dates, names of entities or quantities. *"What is the capital of Brazil?"* is an example of the so called factual questions. In this case, the answer is the name of a city.

On the long road towards the next generation systems, the work presented here takes a new step forward. It defines a layer that, installed on top of current NL-based search engines or QA systems, enhances their capabilities of processing different types of complex temporal questions.

The specific case of temporal QA is not a trivial task due to the potential complexity of temporal questions. Current search engines, such as operational QA systems can deal with simple factual temporal questions, that is, questions requiring a date as an answer (*"When did Bob Marley die?"*) or questions that involve simple temporal expressions in their formulation (*"Who won the U.S. Open in 1999?"*). Processing these kinds of questions is usually accomplished by identifying explicit temporal expressions in questions and the relevant documents that contain these temporal expressions in order to answer the questions.

However, the system described in this paper also processes complex temporal questions. That is, questions whose complexity is related to the temporal properties of the entities enquired about and the relative ordering of events mentioned in the question. The following are examples of these complex temporal questions:

- "Who was the spokesman of the Soviet Embassy in Baghdad *during* the invasion of Kuwait?"
- "Is Bill Clinton *currently* the President of the United States?"

The approach we present in this work tries to imitate the temporal reasoning of a human when solving these types of questions. For example, a person trying to answer the question: *"Who was the spokesman of the Soviet Embassy in Baghdad during the invasion of Kuwait?"* would proceed as follows:

1. First, the complex question would be decomposed into two simple ones: *"Who was the spokesman of the Soviet Embassy in Baghdad?"* and *"When did the invasion of Kuwait occur?"*.

2. He/She would look for all the possible answers to the first simple question: *"Who was spokesman of the Soviet Embassy in Baghdad?"*.

3. After that, he/she would look for the answer to the second question: *"When did the invasion of Kuwait occur?"*

4. Finally, he/she would give as a final answer one of the answers for the first question (if there is any) that have temporal compatibility with the answer to the second question. In this case, the answer to the first question must be temporally compatible with the period of dates associated with *"the invasion of Kuwait"* (*during*).

Therefore, a logical approach for the treatment of complex questions should be based on the decomposition of these questions into simple ones that can be resolved using conventional QA systems. Finally, answers to simple questions, fulfilling the temporal constraints, would be used to construct the answer to the original complex question.

This study presents the development and evaluation of a tool that processes complex NL-temporal questions for information retrieval purposes. Apart from the fact that the tool is capable of processing this type of complex questions, it has the following advantages:

- It can be incorporated as a layer on top of one or more already existing QA systems.
- It can contain and integrate into an answer different data obtained from different types of information sources (web pages, databases, documents, etc.) that are retrieved using different types of search engines (QA, NLIDB[1], etc.).
- The layer is a portable platform since the language-dependent features of the process are easily extended to other languages.
- All the information necessary to process the complex question is obtained directly from it, no extra auxiliary questions or annotations are required.

In this paper, our main aim is to demonstrate how the temporal layer can improve a general purpose QA system when questions are not simple or factual, but of a higher degree of complexity. Specifically, we implemented the temporal layer in order to deal with questions with different levels of temporal complexity. Furthermore, the proposed treatment of questions uses a minimum quantity of linguistic resources in order to obtain a very portable platform, which can be easily extended to different languages.

The paper has been structured in the following way: first of all, section 2 briefly introduces the current situation of temporal reasoning and QA; section 3 depicts our proposal for classifying temporal questions into four groups, depending on the features of the question; section 4 explains the concept of a Multilayered QA system; section 5 describes the different modules of the temporal layer in more detail; and in section 6, decomposition of the question and the Multilayered QA system are evaluated for English. The portability of the system to other languages is then described, and the procedure repeated and evaluated for Spanish. Finally, some conclusions and comments on future work will be made.

1. Natural Language Interfaces to Databases

## 2. Background

As explained in the introduction, one of the aims of this paper is to process complex questions. Complex questions in general have been dealt with in previous studies using different approaches to decompose them. Harabagiu, Lacatusu and Hickl (2006) presented a procedure in which a question produces lots of queries that are semantically related to the original question, with the main aim of obtaining more information about the answers. This approach requires a significant amount of semantic information. The question decomposition presented by Katz, Borchardt and Felshin (2005) involves three decomposition techniques: a) a syntactic decomposition using linguistic knowledge and language-based descriptions of resource content, b) a semantic decomposition using domain-motivated explanation patterns and language-based annotations of resource content, and c) a semantic decomposition of both questions and resource content into lower-level assertions. This approach makes use of a considerable amount of linguistic knowledge and in order to move to new domains, new sets of parameterized language-based annotations need to be composed. In addition to these studies dealing with single focus complex questions, Lin and Lui (2008) propose processing complex questions with multiple foci by obtaining one subquestion for each focus of the original question. This approach determines four possible relations between the subquestions derived from the original question. However, the temporal relation is not considered in this approach.

Apart from complex questions treatment, the motivation for the temporal aspect of this work is due to the great importance in the question answering field of relating questions and information to the temporal dimension in order to find a correct answer. Take, for example, the following two similar questions:

- "Who is the president of the USA?"
- "Who was the president of the USA in 1975?"

There is an obvious dependency of answers on time, so in order to obtain the right answer to these two questions, temporal information needs to be extracted and processed, because the first question refers to the current president of the USA (the exact point in time when the question is formulated), whereas the second one refers to the president in 1975. When the temporal information is not explicit, the questions are considered complex temporal questions.

The importance of the temporal dimension of data in information search processes is corroborated by the recent interest shown by the major evaluation forums on QA like Text REtrieval Conference - TREC (2008) and Cross Language Evaluation Forum - CLEF (2008), see also the works by Voorhees (2002) and Magnini et al. (2005), in including different types of temporal questions as part of their evaluation benchmarks.

Furthermore, CLEF has explicitly fostered research into complex temporal questions by organizing a specific pilot task for such questions (Herrera, Peñas, & Verdejo, 2005) and including in CLEF (Magnini et al., 2006) the temporal dimension of questions and answers as part of its main QA task.

A temporal question can be appropriately processed by: (1) relating the available information to its temporal dimension and (2) adapting the search to link this temporal information with the information search process.

Concerning the first task, the analysis of time is a challenging problem, as the needs of applications based on information extraction techniques expand to include varying degrees of time stamping (identification and reasoning) of events or expressions within a narrative or question. Interest in temporal representation and reasoning has been evolving throughout the years and has resulted in a growing number of meetings related to this topic. We present here, in descending chronological order, the most important ones: TIME (2008) is an annual symposium on Temporal Representation and Reasoning (Demri & Jensen, 2008), it involves different areas including Time in Natural Language; TempEval 2007 (Verhagen et al., 2007) is a workshop held within SemEval-2007 for the evaluation of systems performing Time-Event Temporal Relation Identification; ARTE 2006 is a new workshop focused on Annotating and Reasoning about Time and Events (Ahn, 2006; Dalli & Wilks, 2006; Mani & Wellner, 2006) and was part of the relevant conference COLING-ACL (2006) (Pan, Mulkar, & Hobbs, 2006a); Dagstuhl 2005 was a seminar about annotating, extracting and reasoning time and events (Katz, Pustejovsky, & Schilder, 2005); TERN (2004) was an international competition in which different systems that identify and normalize temporal expressions were evaluated and compared; TANGO 2003 was specialized in developing an appropriate infrastructure for annotation (Pustejovsky & Mani, 2008); LREC (2002) dedicated a workshop to Annotation Standards for Temporal Information in Natural Language (Mani & Wilson, 2002; Setzer & Gaizauskas, 2002; Saquete, Martínez-Barco, & Muñoz, 2002); ACL (2001) included the Temporal and Spatial Information Processing workshop (Setzer & Gaizauskas, 2001; Filatova & Hovy, 2001; Katz & Arosio, 2001; Moia, 2001; Schilder & Habel, 2001; Wilson, Mani, Sundheim, & Ferro, 2001) and finally, COLING (2000), in which some papers were related to temporal expression identification or temporal databases. It is important to emphasize that all these meetings led to the development of a standard for a specification language for events and temporal expressions and their ordering (TimeML, 2008). Nowadays, there is also a growing number of automatic systems extracting temporal expression information[2], such as: ATEL (2008), Chronos (Negri, 2007), TempEx (2008), GUTime (Mani & Wilson, 2000a), DANTE (Mazur & Dale, 2007), TimexTag (Ahn, 2006) and TERSEO (Saquete, Muñoz, & Martínez-Barco, 2006).

Regarding the second task, significant progress has been made in temporal analysis applied to IE and QA tasks as presented in the TERQAS workshop (Pustejovsky, 2002; Radev & Sundheim, 2002). The purpose of the TERQAS workshop was to address the problem of how to enhance natural language question answering systems to answer temporally-based questions about the events and entities in news articles. Besides, a temporal question corpus was developed. As far as we know, one of the first systems that treated temporal information for QA purposes was described by Breck et al. (2000) and it used temporal expression identification applying the temporal tagger developed by Mani and Wilson (2000b). Another important study related to temporal constraints in question answering is presented by Prager, Chu-Carroll and Czuba (2004). They presented a method to improve the accuracy of a QA system by asking auxiliary questions related to the original question whose answers are used to temporally verify and restrict the original answer. This method is called QA-by-Dossier with Constraints and is very suitable for TREC-style factoid questions, but it has the inconvenience of requiring the generation of a set of auxiliary questions. Besides,

2. http://timexportal.wikidot.com/systems

recently, researchers have also focused on other important features in temporal reasoning for final applications, such as: a) event detection: Evita (Saurí, Knippen, Verhagen, & Pustejovsky, 2005) is an application for recognizing events in natural language texts, and this recognition is applied to QA, b) event extension: Pan, Mulkar and Hobbs (2006b) describe a method to automatically learn durations from event descriptions, and c) temporal relations between temporal expressions and events, as described by Lapata and Lascarides (2006).

However, those strategies that implied a complex temporal processing of the question, using only information extracted from the original question and a small amount of linguistic resources for the temporal reasoning were beyond the scope of these investigations.

Our proposal is focused on temporal reasoning for complex temporal questions and so it is necessary to add a new layer to existing systems, thereby allowing these complex questions to be processed (Saquete, Martínez-Barco, Muñoz, & Vicedo, 2004). The decomposition performed by the temporal layer is based only on the temporal relation between the events of the original question, and no other linguistic information is required in the decomposition. In addition, a system that identifies and normalizes temporal expressions was used as a part of the processing layer (Negri, Saquete, Martínez-Barco, & Muñoz, 2006), taking advantage of the multilingual feature of this system in order to use it for cross-lingual tasks.

However, not all the temporal questions need to be treated in the same way since they may have different characteristics, and for this reason, a classification of the different types of temporal questions is also proposed.

## 3. Temporal Questions Taxonomy

Before explaining how to answer temporal questions, they must be classified into different categories since the way to solve them will differ depending on the type of question involved. The temporality of a question depends on two levels of complexity: a) the number of events in the question: Questions formed by a single event and whose answers can be found in a document (simple questions), and questions formed by more than one event that are temporally related and whose answers could be found in multiple documents (complex questions), and b) the temporal information appearing in the question, like implicit or explicit temporal expressions, that needs to be recognized and normalized. The combination of these two features results in four different types of temporal questions.

**Simple Temporal Questions:**

*Type 1: Single event temporal questions without a temporal expression (TE).* These are questions that require a temporal expression as an answer and do not contain any temporal expression in their formulation. These questions are formed by a single event and no temporal reasoning is required, because they are resolved by a QA system directly without a pre or postprocessing of the question. For example: *"When did Jordan close the port of Aqaba to Kuwait?"*. However, since this taxonomy is a temporal question taxonomy, this type of basic temporal questions need to be considered.

*Type 2: Single event temporal questions with a temporal expression.* These are questions that require a temporal reasoning of the temporal expression contained in the formulation of the question. There is a single event in the question, but there are one or more temporal expressions that need to be identified, normalized and annotated. All this temporal infor-mation is necessary to search for the correct answer, due to the fact that it is establishing temporal constraints for the candidate answers. For example: *"Who won the 1988 New Hampshire Republican primary?"*. Temporal Expression: 1988

**Complex Temporal Questions:**

*Type 3: Multiple event temporal questions with a temporal expression.* Questions that contain more than one event, related by a temporal signal. This signal establishes the order between the events in the question. Moreover, there are one or more temporal expressions in the question. These temporal expressions need to be identified, normalized and annotated, and they establish temporal constraints in the answers to the question. For example: *"What did George Bush do after the U.N. Security Council ordered a global embargo on trade with Iraq in August 90?"* In this example, the temporal signal is *"after"* and the temporal constraint is *"between 8/1/1990 and 8/31/1990"*. This question consists of these two events:

- Event 1: George Bush did *something*
- Event 2: the U.N. Security Council ordered a global embargo on trade with Iraq (Temporal constraint: *"August 1990"*)

*Type 4: Multiple event temporal questions without a temporal expression.* Like Type 3, these questions consist of more than one event, related by a temporal signal, but in this case, the questions do not contain temporal expressions. The temporal signal establishes the order between the events in the question. For example: *"Who was the president of US when the AARP was founded?"*. In this example, the temporal signal is *when* and the question would be decomposed into:

- Event 1: *someone* was the president of US
- Event 2: the AARP foundation

How to process each type of question will be explained in detail in the following sections.

## 4. Architecture of a Multilayered QA System

In order to process special types of questions which are beyond the scope of currently QA systems, this work proposes a multilayered architecture that increases the functionality of these QA systems, allowing them to solve any type of complex question. In this work, the temporal layer has been implemented. Moreover, this architecture enables different layers to be added to cope with questions that need other kinds of complex processing and are not temporally oriented, such as script questions (*"How do I assemble a bicycle?"*) or template-based questions (*"What is the main biographical data of Nelson Mandela?"*).

Complex questions have in common the need for additional processing of the question in order to solve it adequately. The architecture presented in this paper enables different types of complex questions to be dealt with by superposing additional processing layers, one for each type, on the top of an existing general purpose QA system, as shown in Figure 1. These layers will:

- decompose the question into simple events to generate simple questions (sub-questions) that are ordered according to the original question,

- send simple questions to a general purpose QA system,
- receive the answers to the simple questions from the general purpose QA system,
- filter, compare and validate the sub-answers, according to the relation detected between sub-questions, in order to construct the final complex answer.

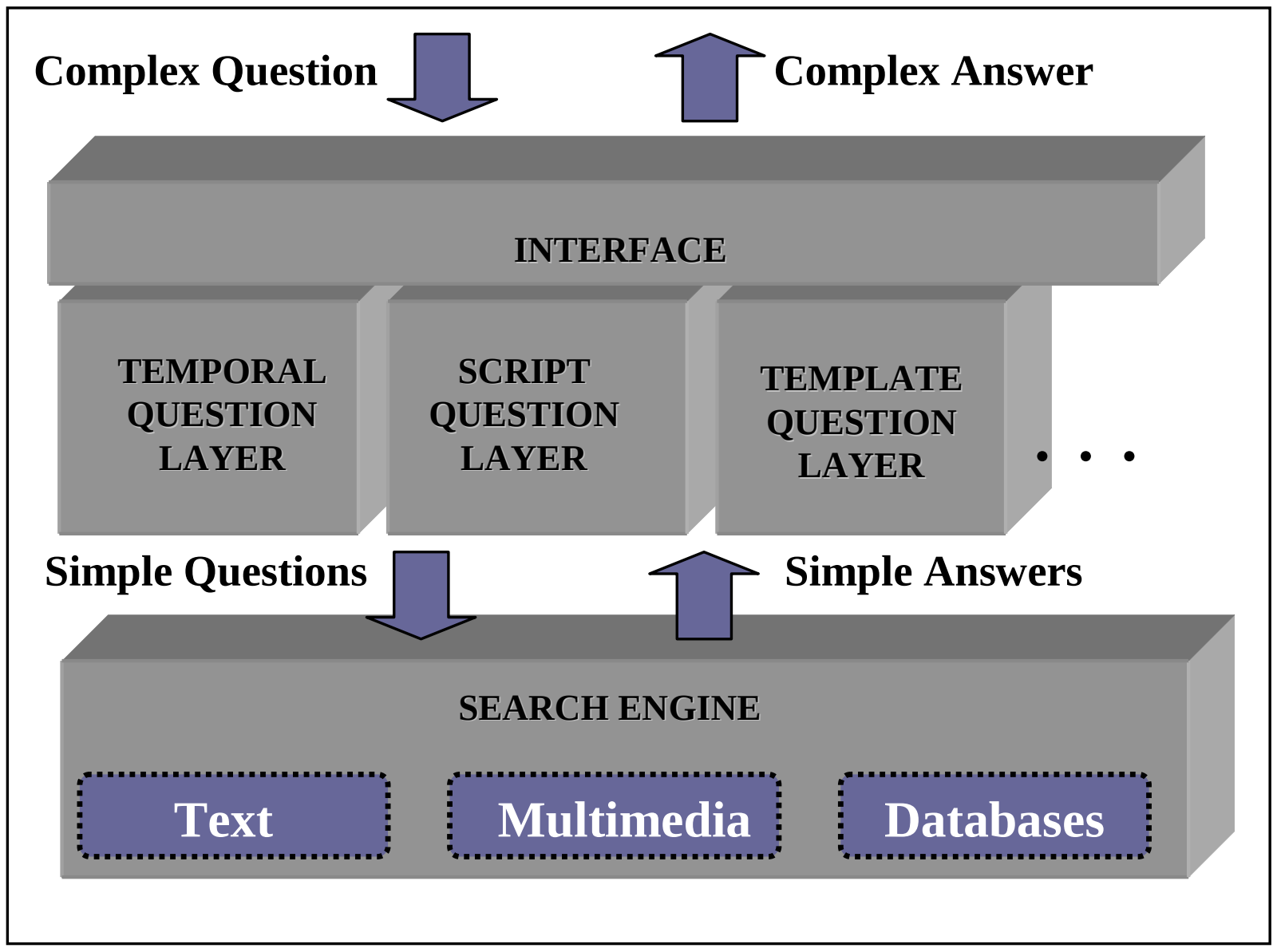

Figure 1: Multi-layered Architecture of a QA system

This architecture has a large number of advantages, of which the following should be mentioned:

- It allows researchers to use any existing general purpose QA system.
- Since complex questions are processed by a superior layer, it is not necessary to modify the current QA system when you want to deal with more complex questions. The layer enhances the capabilities of an existing QA system without changing it in any way.
- Each additional processing layer works independently from the others and only processes the questions accepted by that layer.
- It is possible to have more than one type of QA system working in parallel, each of them specialized in searching for a specific type of information (text,multimedia,databases).

Next, a layer oriented to processing temporal questions according to the taxonomy shown in section 3 is presented.

## 5. Temporal Layer

The temporal layer proposed here consists of two units, the *Question Decomposition Unit* and the *Answer Recomposition Unit*, which will be superposed over a general purpose QA system, as shown in Figure 2.

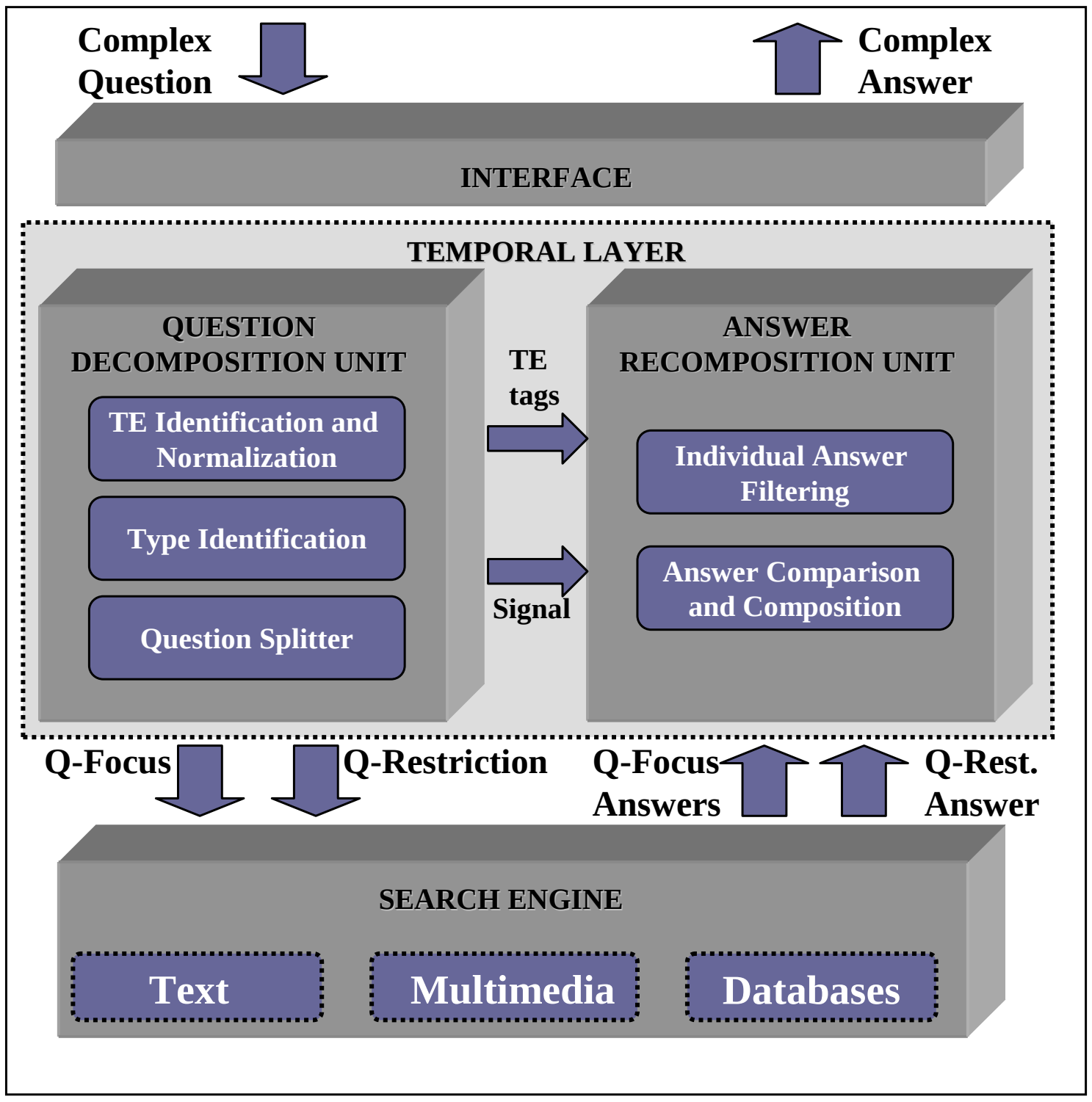

Figure 2: Architecture of the temporal layer

These components all work together in order to obtain a final answer as follows:

- *Question Decomposition Unit* is a preprocessing unit which performs three main tasks. First of all, temporal expressions in the question are identified and normalized. Secondly, following the taxonomy shown in section 3, there are different types of questions and each type must be treated in a different way. For this reason, the type needs to be identified. After that, complex questions (Type 3 and 4) are split into simple ones using the temporal signal as a reference. The first sub-question is defined as the question focus (Q-Focus) and it specifies the type of information the user needs to find. The second sub-question is called the question restriction (Q-Restriction) and the answer to this sub-question establishes the temporal restrictions on the list of answers to the Q-Focus. The Q-Focus and the Q-Restriction are the input of the QA system. For example, the question *"Where did Bill Clinton study before going to Oxford University?"*, is divided into two sub-questions that are related by the temporal signal *"before"*: Q-Focus: *"Where did Bill Clinton study?"* and Q-Restriction: *"When did* Bill Clinton *go to Oxford University?"*.

- *General purpose QA system.* The simple questions generated are processed by a general purpose QA system. Any QA system could be used here (QA systems, Multimedia search engines or NLIDB). For the example above, a current QA system returns the following answers:
  - Q-Focus Answers:
    - Georgetown University (1964-68)
    - Oxford University (1968-70)
    - Yale Law School (1970-73)
  - Q-Restriction Answer: 1968
- *Answer Recomposition Unit.* This unit constructs the answer to the original question from the answers to the Q-Focus and the Q-Restriction using all the temporal constraints, such as temporal signals (which are fully explained later) or temporal expressions, available in the original question. The temporal signal establishes the appropriate order between the answers to the Q-Focus and the Q-Restriction in the question. Finally, this unit returns the appropriate answer by analyzing the temporal compatibility between the list of possible Q-Focus answers and the Q-Restriction answer.

An example of how the temporal layer operates is shown in Figure 3.

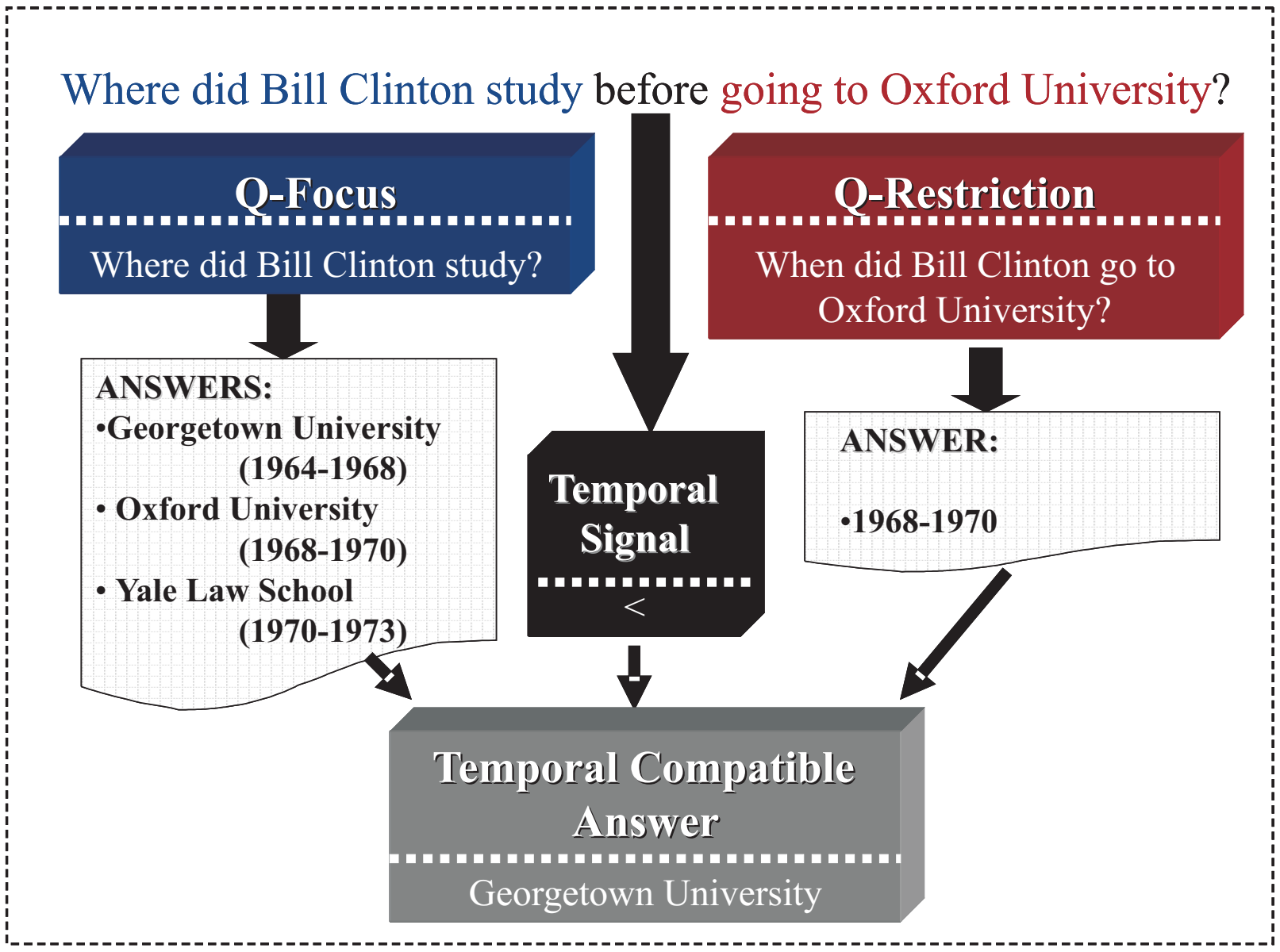

Figure 3: Example of performance of the Temporal Layer

It is important to emphasize that the temporal layer is a language dependent platform (it uses lexical and syntactic patterns) and English was the language chosen initially to develop the layer; however, it can be easily extended to other languages, as will be seen in section 6.3. The units that integrate the temporal layer are described in more detail in the following sections.

### 5.1 Question Decomposition Unit

The main task of this unit, which is divided into three main modules, is the temporal reasoning of the temporal information of the question and the decomposition of the question (only required in Type 3 and 4 questions). The *temporal expression identification and normalization* module detects and resolves the temporal expressions in the question. The *type identification* module classifies the question according to the taxonomy proposed in section 3. Finally, the *question splitter* module splits the complex question into simple ones.

Thus, the output of the Question Decomposition unit consists of:

- two sub-questions (Q-Focus and Q-Restriction), which will be processed by a QA system in order to obtain an answer for each of them,
- temporal tags, containing concrete dates returned by the TERSEO system (Saquete et al., 2006), these tags are part of the input of the Answer Recomposition Unit and they are used by this unit as temporal constraints in order to filter the individual answers, and
- the temporal signal, which is part of the input of the Answer Recomposition Unit as well, because this information is needed in order to compose the final answer and determine the temporal compatibility between the answers to the Q-Focus and the answer to the Q-Restriction.

The modules of the decomposition unit are fully explained in the following subsections.

#### 5.1.1 Temporal Expression Identification and Normalization

This module uses the TERSEO system (Saquete et al., 2006) to identify, annotate and normalize temporal expressions in the question.

With this system, implicit and explicit temporal expressions can be annotated. Expressions like “12/06/1975” are explicit, while those like “two days before” are implicit and need the location of another complete temporal expression (TE) to be understood. For the specific purposes of the temporal layer, TERSEO simply returns the text of the temporal expression in a string and the normalization or resolution value of the temporal expression using the ISO standard format for concrete dates or periods.

In this work, TERSEO does not use a complete text as input, but only a question. The temporal tags (`TE` tag with a `value` attribute) obtained from the questions are the output of this module and they are used in the Answer Recomposition Unit in order to filter the individual answers obtained by the QA system. The `TE` tag is necessary in order to determine the temporal compatibility between the answers to the Q-Focus and the answer to the Q-Restriction. For example, in a question like: *“Which U.S. ship was attacked by Israeli forces during the Six Day war in the sixties?”*, the temporal constraint that must be fulfilled is: “the date of Q-Focus answers should be between *1960-01-01* and *1969-12-31*” (“196” in ISO format). This means that only the answers whose dates are within the range of dates in the question are temporally compatible.

It is very important to emphasize that, initially, the TERSEO was developed for Spanish, but a platform to automatically extend the system to other languages was developed as well. Therefore, the system was evaluated for three different languages: Spanish, English and Italian. For Spanish the results were 91% precision and 73% recall. The system was evaluated for English using the TERN (2004) corpus and the results obtained for the F-measure were 86% for identification and about 65% for normalization. For the Italian evaluation, the I-CAB corpus was used. This corpus consists of 525 news documents taken from the local newspaper L'Adige [3]. Ita-TERSEO obtained an F-measure of around 77% for identification. The results are quite good because the extension to English and Italian was completely automatic and therefore, also very fast.

The multilingual capabilities of TERSEO are very interesting in various NLP fields, in particular its application to Crosslingual QA systems, and therefore in the Temporal Layer as well.

#### 5.1.2 Type Identification

The Type Identification module classifies the question into one of the four types in the taxonomy proposed above. This identification is necessary because each type of question produces a different behavior (scenario) in the system. Type 1 and Type 2 questions are classified as simple, and the answer can be obtained without splitting the original question. On the other hand, Type 3 and Type 4 questions need to be split into a set of simple sub-questions. These types of sub-questions are always Type 1, Type 2 or a non-temporal question, which are considered simple questions.

The question type is established according to the rules in Figure 4. There are four possibilities: (a) if there is no Temporal Expression and no Temporal Signal, the question is classified as *Type 1*; (b) if there is no Temporal Expression but a Temporal Signal, the question is classified as *Type 4*; (c) if there is a Temporal Expression but no Temporal Signal, the question is classified as *Type 2*; (d) if there is a Temporal Expression and a Temporal signal, the question is classified as *Type 3*.

#### 5.1.3 Question Splitter

This task is only necessary when, according to the type identification module, the question is Type 3 or Type 4. These questions are considered complex questions and need to be divided into simple ones (Type 1, Type 2 or non-temporal questions). The decomposition of a complex question is based on the identification of temporal signals, which link simple events to form complex questions (see Table 1).

As explained before, using the temporal signal as a referent, the two events related by it will be transformed into two simple questions: Question-Focus (Q-Focus) and Question-Restriction (Q-Restriction).

The Q-Focus is a question that specifies the information that the user is searching for. This question is very simple to obtain, because no syntactic changes are required to construct it, only the question mark must be added. When the Q-Focus is processed by a QA system, the system will return a list of possible answers.

The Q-Restriction is constructed using the part of the complex question that follows the temporal signal. This question is always transformed to a *"When"* question using a set

3. http://www.adige.it

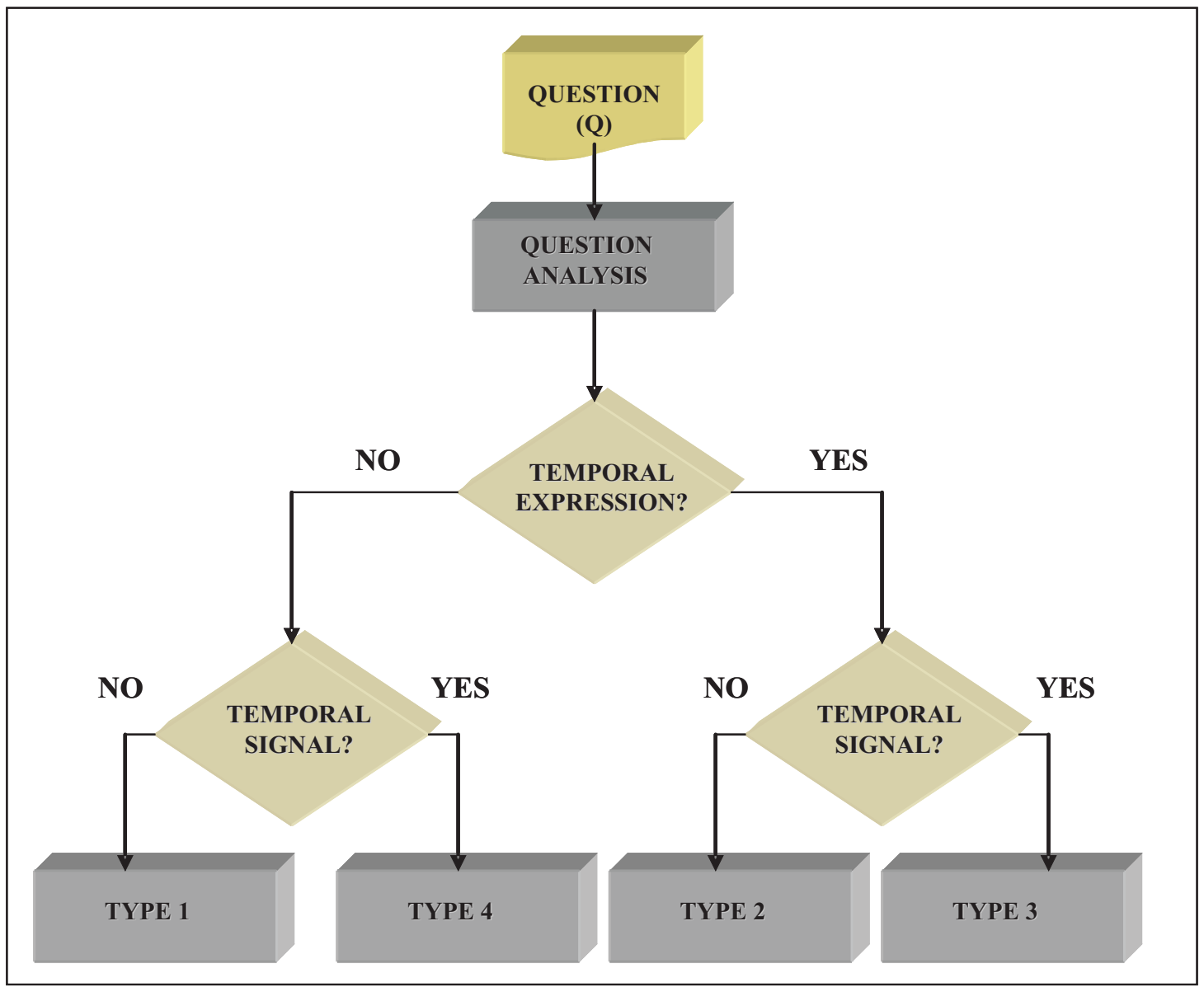

Figure 4: Decision tree for Type Identification

of lexical and syntactic patterns defined in the layer. When the Q-Restriction is processed by a QA system, only one appropriate answer is expected.

In addition, temporal signals denote an ordering between the events being linked. Assuming that *F1* is the date associated with the answers to the Q-Focus and *F2* is the date associated with the answer to the Q-Restriction[4], the signal will establish a certain order between the answers, which is called the *ordering key*. An example of some ordering keys are shown in Table 1.

Table 1: Example of signals and ordering keys

| **SIGNAL** | **ORDERING KEY** |
|---|---|
| After | F1 > F2 |
| When | F1 = F2 |
| Before | F1 < F2 |
| During | F2i <= F1 <= F2f |
| From F2 to F3 | F2 <= F1 <= F3 |
| About F2 – F3 | F2 <= F1 <= F3 |
| On / in | F1 = F2 |
| While | F2i <= F1 <= F2f |
| For | F2i <= F1 <= F2f |
| At the time of | F1 = F2 |
| Since | F1 > F2 |

Using the list of answers to the Q-Focus, the answer to the Q-Restriction and the temporal signal, the Answer Recomposition Unit determines the temporal compatibility

4. F2:Q-Restriction concrete date / [F2i-F2f]:Q-Restriction period dates

between the answers and composes the final answer to the original complex question. This process will be fully explained in the following subsection.

### 5.2 Answer Recomposition Unit

The main task of the Answer Recomposition Unit is to obtain the final answer to the complex question using all the available inputs of the Decomposition Unit and the answers obtained from the QA system. The Recomposition Unit is divided into two modules. The *Individual Answer Filtering* module filters the possible answers to the Q-Focus, avoiding the non-temporally compatible ones, and the *Answer Comparison and Composition* module, which composes the answer to the original question by using the ordering key established by the temporal signal.

Once the complex questions have been split by the Decomposition Unit into the Q-Focus and the Q-Restriction and the answers to these questions have been obtained by a QA system, the Recomposition Unit determines from the list of all the potential answers to the Q-Focus which one is compatible with all the temporal constraints obtained in the process: temporal expressions, temporal signal and answer to the Q-Restriction. The answers to the Q-Focus that fulfill the temporal constraints will be considered the answer to the initial complex question.

#### 5.2.1 Individual Answer Filtering

The list of possible answers to the Q-Focus and the answer to the Q-Restriction given by the QA system are the input of the Individual Answer Filtering module. For a Q-Focus or Q-Restriction with a temporal expression, it selects only those answers that satisfy the temporal constraints obtained by the TE Identification and Normalization Unit. The date of the answer should be temporally compatible with the temporal tag, that is, the date of the answer must lie within the date values of the tag. If not, it will be rejected. Only the answers that fulfill the constraints go to the Answer Comparison and Composition module.

#### 5.2.2 Answer Comparison and Composition

Finally, once the answers have been filtered using the signals and the ordering key, the results for the Q-Focus are compared with the answer to the Q-Restriction in order to determine if they are temporally compatible. Temporal signals denote the relationship of order between the date of the answer to the Q-Focus and the date of the answer to the Q-Restriction.

Only the answers that fulfill the compatibility established by the temporal signal can be possible answers to the original question. The answer selected is considered by this module to be the answer to the complex question. Hence, the system is able to solve complex temporal questions.

## 6. Evaluation Experiments

The evaluation experiments performed in this paper were done initially for English, and after porting the system to Spanish, the same evaluation procedure was carried out for this new language.

This evaluation has a dual aim: on the one hand, to determine if the Decomposition Unit is able to process each type of question properly in order to obtain appropriate simple factual questions that can be answered by any kind of general purpose QA system, and on the other hand, to show to what extent the general purpose QA system could be improved when these techniques are applied.

### 6.1 Test Environment

First of all, a corpus of English questions that contains as many simple and complex temporal questions as possible was necessary. The first idea was to use already existing resources, such as TREC (2008) and CLEF (2008) question corpora, due to the large number of questions they contain. Unfortunately, after studying these corpora, they were discarded because they do not contain complex temporal questions. Thus, using the initial TERQAS question corpus proposal (Radev & Sundheim, 2002; Pustejovsky, 2002) as a model, a new question corpus was manually developed collecting questions from a group of volunteers unacquainted with this work. The instructions given to the volunteers were: 1) answers to the questions proposed must be found on the Internet, 2) the questions must be constructed according to the temporal question taxonomy described in Section 3, and 3) the questions must expect a fact as an answer (factual questions). In the case of complex questions, two factual questions must be related to a temporal signal[5]. This last instruction was necessary in order to make the evaluation procedure more straightforward since open-ended questions usually require long answers, which makes them more difficult to judge. In order to have a balanced corpus, some questions were discarded and finally the corpus developed contains a balanced number (50) of each type of temporal question (Types 1,2,3 and 4), which resulted in 200 temporal questions for English[6]. For the Spanish evaluation, the English question corpus was manually translated into this language. Therefore, the distribution of the questions by type is the same as in English.

Once the question corpus for English and Spanish were developed, the following step was to construct the testbeds for both languages in order to allow for rigorous, transparent and replicable evaluation tests. The testbed annotation was performed using an XML schema and was developed by three independent annotators. In the case of doubts or disagreement, the annotation was reviewed by a referee, who made the final decision. The interannotator agreement was calculated for every attribute, resulting 100% for all cases except the temporal signal (98%) due to the complexity of some temporal signals.

In the testbed annotation, every question is annotated with a `Q` tag, and this tag has an `id` attribute that identifies every question. The question string is annotated using the `QUESTION` tag. Furthermore, for every question, there are five items that must be annotated:

1. *Identification and Normalization of the temporal expressions* in the question. The annotation for this item is done using a `TE` tag, and its content stores the string text of the TE. The tag also has an attribute `value`, which stores the resolution of the expression using the ISO format,

5. It is important to emphasize that Type 3 questions contain two events and a temporal expression, which is used as an extra time constraint in the answering procedure, limiting the number of potential answers, speeding the answering step, and refining the final answer
6. http://gplsi.dlsi.ua.es/corpus/CTQ

2. the *type of question* according to the classification proposed in this paper. This type is annotated using the tag `TYPE` and it must be a value between 1 and 4. Since the questions were manually built using the temporal question taxonomy as a reference,

3. the *temporal signal*. The tag used to annotate this item is called `SIGNAL` and it stores the exact string text of the temporal signal of the question,

4. the *two possible sub-questions* in the case of complex questions (Type 3 and 4): the first sub-question is annotated using the tag `Q-FOCUS` and the second using the tag `Q-REST`, and,

5. the *answer* to the complex question: the answer is annotated using an `ANSWER` tag and contains the correct answer to the complex question. One example of the annotation format for a question is:

```
<Q id="107">
    <QUESTION>Who won the best actress Oscar award when James Dean died in the 50s?< /QUESTION>
    <TE value="195">the 50s< /TE>
    <TYPE>3< /TYPE>
    <SIGNAL>when< /SIGNAL>
    <Q-FOCUS>Who won the best actress Oscar award?< /Q-FOCUS>
    <Q-REST>When did James Dean die in the 1950s?< /Q-REST>
    <ANSWER>Anna Magnani< /ANSWER>
< /Q>
```

In the case of the Decomposition Unit, the following five aspects are evaluated:

- *TE Identification and Normalization*: are the temporal expressions in the question correctly detected and normalized?

- *Type Identification*: is the type of question correctly identified according to the classification presented previously in this paper?

- *Signal Detection*: are any temporal signals in the question correctly detected?

- *Question Splitter*: are the complex questions correctly split into simple factual questions, which can be answered by a general purpose QA system?

- *DECOMPOSITION UNIT* as a whole: has the system correctly undertaken all the sub-tasks previously defined, since these sub-tasks as a whole compose the decomposition unit?

However, not all the evaluation aspects explained above need to be considered for all the questions. Table 2 determines if an aspect must be evaluated or ignored for each particular type of question. The decomposition unit as a whole is only taking into consideration the evaluated sub-tasks for every question.

Having determined *what* aspects are evaluated in the decomposition and *in what cases* they must be evaluated, depending on the type of question, the next step is to establish *how* these aspects are evaluated. For this purpose, the criteria matrix, containing the rules followed in the evaluation in order to determine when the elements are treated (ACT) and which of them are correct (CORR), is shown in Table 3.

Table 2: Aspects evaluated for Decomposition Unit depending on the Q-type

| **Type** | **TE Id.Norm.** | **Type** | **Signal** | **Q Splitter** | **DECOMP.** |
|---|---|---|---|---|---|
| 1 | – | Yes | – | – | Yes |
| 2 | Yes | Yes | – | – | Yes |
| 3 | Yes | Yes | Yes | Yes | Yes |
| 4 | – | Yes | Yes | Yes | Yes |

Table 3: ACT and CORR criteria matrix for Decomposition Unit evaluation

| | **ACT** | **CORR** |
|---|---|---|
| **TE Ident.Norm.** | The TE is annotated by the system | -Exact agreement with `TE` tag content<br>-Exact agreement with `value` attribute content within `TE` tag |
| **Type** | The type is returned by the system | Exact agreement with `TYPE` tag content |
| **Signal** | The temporal signal is annotated by the system | Exact agreement with `SIGNAL` tag content |
| **Q Splitter** | Complex Q is divided into two sub-Q's | Sub-Q's agreement of `Q-FOCUS` and `Q-REST` tag in:<br>-Interrogative particle<br>-Main verb correctly detected and tensed<br>-All keywords appear and only keywords from the original Q, except stopwords |
| **DECOMP.** | All previous aspects ACT | All previous aspect CORR |

In the case of QA evaluation, we are using the current **CLEF evaluation criteria**[7] as starting point, determining correct and inexact answers. The use of these evaluation criteria is possible since our corpus only contains factual questions. Therefore, determining the correctness of an answer is very straightforward. The other CLEF judgments are not specified in this evaluation because the measure of incorrectness may be calculated directly by subtracting the number of correct answers from the total number of questions. In addition, the unknown judgement is also omitted since two human assessors must evaluate all the answers. And finally, we do not consider unsupported judgement neither, since our corpus consisted of data obtained from the Internet, where all correct answers can be found.

The criteria matrix for QA, shown in Table 4, describes the rules followed in the evaluation in order to determine treated (ACT), correct (CORR) and inexact (INE) answers.

Table 4: ACT and CORR criteria matrix for QA system

| | **ACT** | **CORR (CLEF R)** | **INE (CLEF X)** |
|---|---|---|---|
| **QA** | An anwer is returned by the system | -Exact agreement with one of the answers contained in `ANSWER` tag content | The answer contains a correct answer, but it is incomplete or longer than the minimum amount of information required |

For all the evaluations performed in this work, the following measures were used:

- POS:Total number of items

7. http://www.clef-campaign.org/

- ACT: Number of items treated by the system
- CORR: Number of items properly treated (Correct)(CLEF R)
- PREC: Precision ($\frac{CORR}{ACT}$) percentage of items in the output of the system that are properly treated
- REC: Recall ($\frac{CORR}{POS}$) percentage of items treated by the system (CLEF Accuracy)
- F: ($\frac{(1+\beta^2)(P*R)}{(\beta^2*P+R)}$) Combination of precision and recall in a single value. $\beta = 1$
- Only for QA evaluation purposes:
  - MRR: For an ordered list of possible answers of a question is ($\frac{1}{CorrectAnswerPosition}$). The final MRR is the average of every individual MRR obtained.
  - INE: Is the number of answers judged inexact by human assessors (CLEF X).

### 6.2 Evaluation Results

This section presents the results of the decomposition unit and an analysis of its influence in QA systems for English.

#### 6.2.1 Evaluating the Decomposition Unit for English

In this section, the decomposition unit[8] for the processing of simple and complex temporal questions in English is evaluated, based on the testbed defined previously. In this evaluation, in addition to the decomposition unit efficiency, some aspects of temporal expressions and their influence on complex questions were analyzed.

The evaluation results are very good with an F-measure of 89.5%. All the results are shown in Table 5. In the evaluation, 176 of a total of 200 questions were correctly preprocessed. Since the decomposition unit not only divides the complex questions but also determines the type of the question and performs the temporal reasoning if necessary, the whole set of questions is considered in the global measure of the decomposition unit. It is obvious that in the case of Type 1 questions the decomposition unit simply determines the type of question, but we were interested in evaluating the performance of the unit in this respect.

Table 5: Evaluation of the decomposition unit for English

| | POS | ACT | CORR | PREC | REC | F |
|---|---|---|---|---|---|---|
| **TE Identification and Normalization** | 100 | 93 | 80 | 86.0% | 80.0% | 82.9% |
| **Type Identification** | 200 | 200 | 194 | 97.0% | 97.0% | 97.0% |
| **Signal Detection** | 100 | 100 | 96 | 96.0% | 96.0% | 96.0% |
| **Question Splitter** | 100 | 100 | 92 | 92.0% | 92.0% | 92.0% |
| **DECOMPOSITION UNIT** | **200** | **193** | **176** | **91.1%** | **88.0%** | **89.5%** |

8. http://gplsi.dlsi.ua.es/demos/TQA/

Next, a detailed analysis of the results for each evaluation aspect is shown (see Appendix A for detailed error examples):

- *Identification and normalization of Temporal Expressions*: In this corpus, there were 100 temporal expressions and our system detected 93, of which 80 were correctly resolved. As we said previously, this module uses the TERSEO system to identify and normalize the temporal expressions in the question. There were three types of errors: (1) expressions that were treated by TERSEO as temporal expressions but were not in fact temporal; (2) expressions that were identified wrongly because: a) the expression is outside the scope of the TERSEO system, or b) the identification extent is not exact; and (3) expressions that: a) were normalized wrongly because the normalization rule in TERSEO was not appropriate for these expressions, or b) were not normalized because the normalization rule does not exist.

- *Type Identification:* There were 200 temporal questions in the corpus, all of them were processed by this module, and 194 were correctly identified according to the taxonomy proposed in section 3. The errors in this module were due to the fact that some TE were not detected by TERSEO, as shown in Appendix A. However, this type of error does not usually affect the question splitting and in most cases the complex question is split correctly.

- *Signal Detection:* In the corpus, there were 100 questions that were considered complex (Type 3 and Type 4). Our system was able to treat and recognize correctly the temporal signal of 96 of these questions. The main error detected in this module arose when a temporal expression was part of a signal, denoting a complex signal, such as: *"EVENT1 a year after EVENT2"*. This type of complex signal is outside the scope of the system. The system also fails when a preposition, classified as a signal in the system, is part of a TE, like *"during the 18 century"* and therefore *"during"* is wrongly considered a temporal signal.

- *Question Splitter:* From this set of 100 complex questions, the system was able to process and split 92 of them properly. The errors in this unit are due to: a) wrong signal identification; or b) syntactic problems, obtaining a tensed verb or the subject of the Q-Focus to construct the Q-Restriction properly. For instance, in the question *"Which language was invented when Berliner patented the Gramophone?"*, the POSTagger did not identify *"patented"* as a past tense verb and the Q-Restriction was wrongly generated as: *"When did Berliner patented the Gramophone happen?"*.

One possible problem that can appear in complex questions, and is not yet treated by our system, are questions that contain anaphoric co-references. Therefore, when splitting the complex question into two separate simple questions, the question that contains the anaphoric co-reference can not be treated directly by a QA system and needs to be processed by a module that performs anaphora resolution first. For example: *"In which studies did Ms. Whitman graduate before she got her MBA? "*. The Q-Restriction obtained is: *"When did she get her MBA?"*. In this case, *"she"* is referring to *"Ms. Whitman"*. In our case, this type of question is outside our scope. However, just by adding a module that adapts anaphora resolution techniques for dialogs and texts (Palomar et al., 2001; Palomar &

Martínez-Barco, 2001) to questions, the situation will be solved. Moreover, applying this module to the question does not affect the decomposition process in any case.

#### 6.2.2 Evaluating the influence of the temporal processing in QA systems for English

The QA system used for this evaluation is a general purpose one that uses Internet data as the corpus (Moreda, Llorens, Saquete, & Palomar, 2008a). This is a very simple open-domain QA system, whose main feature is that the Answer Extraction Unit is able to look for possible answers in two ways: performing a mapping with the type of name entity that the question requires (NE-based), or with the type of semantic role that the question needs as an answer (SR-based)[9].

Due to the modularity of the QA system, in this evaluation, only the NE-based answer extraction is used. As a baseline, using a subset of factual questions, extracted from TREC1999 and TREC2000 that are NE oriented, the authors evaluated the system and found 87.50% precision, 84% recall, 85.70% F and 87.25% MRR (Moreda, Llorens, Saquete, & Palomar, 2008b).

The evaluation performed in this work is divided into two experiments:

1. Base QA system evaluation: First the QA system is evaluated without using the temporal layer.

2. Multilayered QA system evaluation: Then the QA system is evaluated when it performs with the temporal layer.

The main aim of this evaluation is to compare the results of the two experiments and determine if the temporal layer enhances a general purpose QA system like the one used in this case. Besides, for both experiments, the 200 temporal question corpus created for this purpose containing simple (Type 1 and Type 2) and complex (Type 3 and Type 4) questions is used.

The results obtained by the general purpose QA system without the temporal layer are shown in Table 6.

Table 6: Evaluation of the QA system for English temporal questions

| | POS | ACT | CORR | INE | PREC | REC | F | MRR |
|---|---|---|---|---|---|---|---|---|
| **Type 1** | 50 | 50 | 35 | 0 | 70.00% | 70.00% | 70.00% | 77.60% |
| **Type 2** | 50 | 45 | 23 | 1 | 51.11% | 46.00% | 48.42% | 48.00% |
| **Type 3** | 50 | 8 | 1 | 0 | 12.50% | 2.00% | 3.45% | 3.00% |
| **Type 4** | 50 | 18 | 2 | 0 | 11.11% | 4.00% | 5.88% | 5.00% |
| **GLOBAL** | 200 | 121 | 32 | 1 | 50.41% | **30.50%** | 38.01% | 33.40% |

The results obtained by the system enhanced with the temporal layer are shown in Table 7.

9. http://gplsi.dlsi.ua.es/demos/TMQA/

Table 7: Evaluation of QA system plus temporal layer for English temporal questions

| | POS | ACT | CORR | INE | PREC | REC | F | MRR |
|---|---|---|---|---|---|---|---|---|
| **Type 1** | 50 | 50 | 35 | 0 | 70.00% | 70.00% | 70.00% | 77.60% |
| **Type 2** | 50 | 47 | 38 | 1 | 80.85% | 76.00% | 78.35% | 78.00% |
| **Type 3** | 50 | 48 | 29 | 2 | 60.42% | 58.00% | 59.18% | 63.66% |
| **Type 4** | 50 | 46 | 26 | 2 | 56.52% | 52.00% | 54.17% | 55.66% |
| **GLOBAL** | 200 | 191 | 128 | 5 | 67.02% | **64.00%** | 65.47% | 68.73% |

As shown in both tables, the QA system enhanced with the temporal layer offers better results in all measures (72.24% improvement in F and 33.58% error reduction in F). The most outstanding improvements occur in complex temporal questions, due to the extra reasoning that the temporal layer applies to find a candidate answer. Moreover, an extra experiment, with manually corrected temporal expression identification and normalization, is performed. Obviously, only questions Type 3 and 4 are affected and improved. Results are shown in Table 8.

Table 8: Evaluation of QA system plus temporal layer for English temporal questions with manually corrected TERN

| | POS | ACT | CORR | INE | PREC | REC | F | MRR |
|---|---|---|---|---|---|---|---|---|
| **Type 1** | 50 | 50 | 35 | 0 | 70.00% | 70.00% | 70.00% | 77.60% |
| **Type 2** | 50 | 48 | 40 | 1 | **83.33%** | **80.00%** | **81.63%** | **82.00%** |
| **Type 3** | 50 | 48 | 30 | 2 | **62.50%** | **60.00%** | **61.22%** | **65.66%** |
| **Type 4** | 50 | 46 | 26 | 2 | 56.52% | 52.00% | 54.17% | 55.66% |
| **GLOBAL** | 200 | 192 | 131 | 5 | **68.22%** | **65.50%** | **66.83%** | **70.23%** |

A graphical comparison of the results for each type of question is shown in Figure 5. It is very clear that the Multilayered QA system enhances the performance of the QA system in all the types of questions except Type 1 (simple factual temporal questions), since this type of question is processed in the same way by both systems. For the other types, precision, recall, F-measure and MRR are improved, especially in the case of Type 3 and Type 4 questions, in which the base QA system is almost incapable of answering these questions properly. The system gave very few inexact answers since the questions need short answers consisting only of an NE or TE.

Some interesting examples that have been analyzed are shown in Figure 6.

In the first example, the question is a Type 2 question, which contains an implicit temporal expression *"16 years ago"*. The question is processed by both systems, but with the important difference that the Multilayered QA system is able to process the temporal expression and normalize the expression to a concrete date, in this case *"1992"*. Once this preprocessing of the temporal expression is done, the question is processed by the Base QA system as *"Where were the Olympics held in 1992?"*, allowing the system to find the correct answer. Without this preprocessing of the temporal layer, the Base QA system returns the most popular answer, which corresponds to the last Olympic games in Beijing, and therefore fails to answer the question correctly.

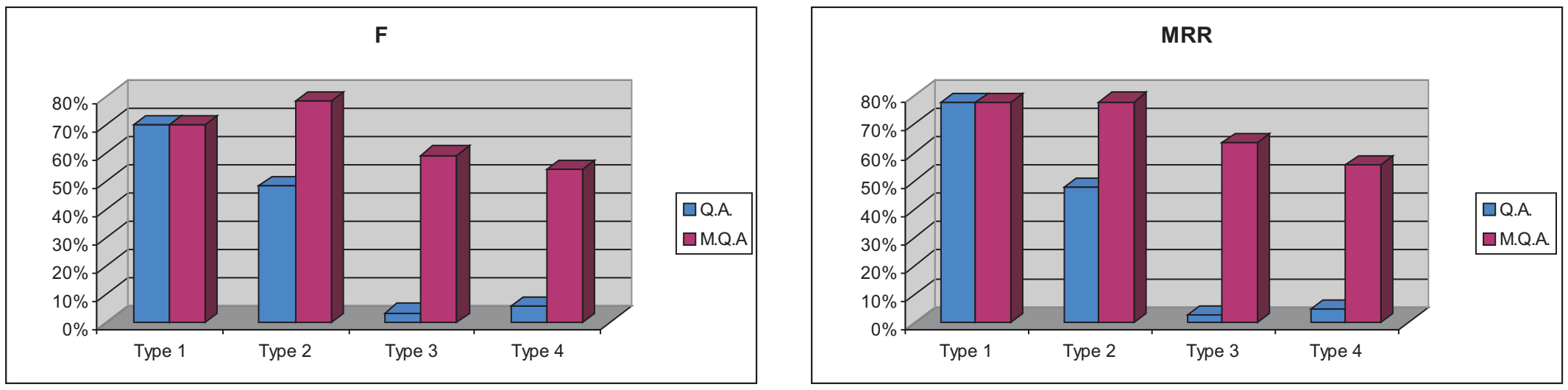

(a) *F-measure comparison* (b) *MRR comparison*

Figure 5: *Comparative graphics between Base QA system and Multilayered QA system*

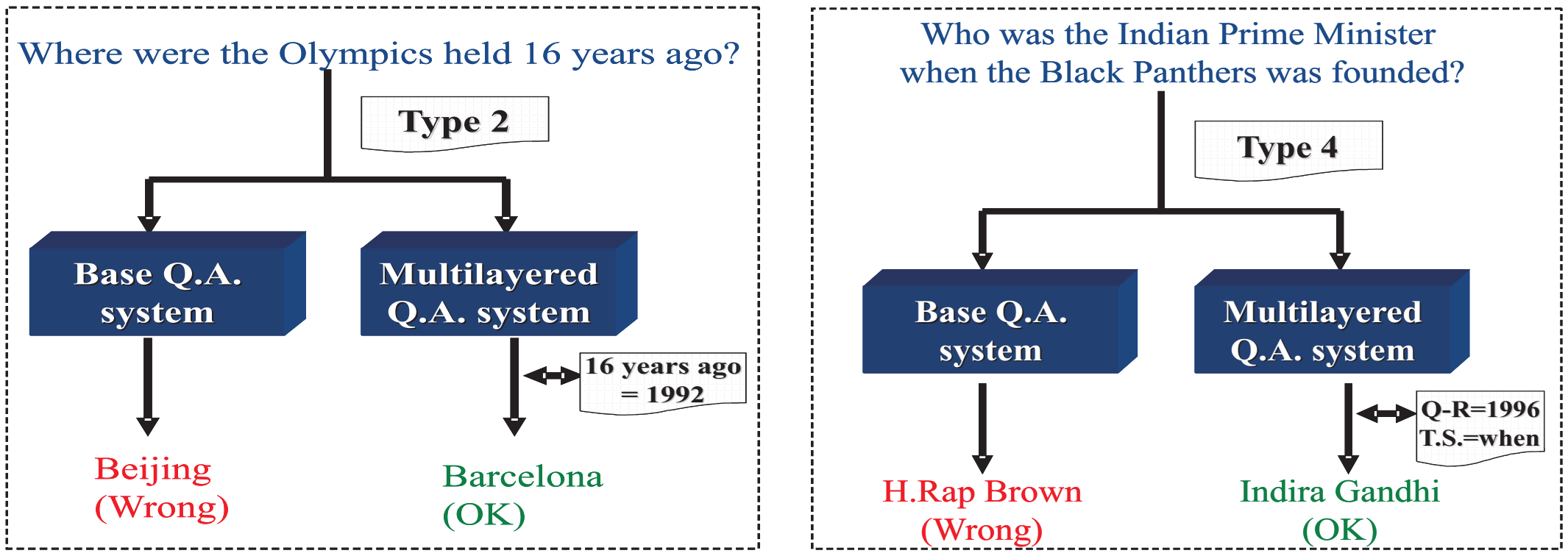

(a) *Example 1* (b) *Example 2*

Figure 6: *Examples of Multilayered QA system performance*

In the second example, the question is a Type 4 complex question and is again processed by both systems. Since the Base QA system is not able to reason the second part of the question and simply uses the keywords in the question, only the Multilayered QA system returns a correct answer, taking as a restriction the date when the event in the second part of the questions occurred.

To conclude, this study demonstrates that including this type of layer can help general purpose QA systems to resolve questions that are more complex than simple factual questions, without changing the implementation of the general QA system.

### 6.2.3 Comparison with other QA systems

In order to compare our results with those of another QA system, we carried out the above test with the widely known START QA system (Katz, 1990, 1997), which is available on the Internet[10]. The results obtained with the START system and those obtained by our

10. http://start.csail.mit.edu/

QA system enhanced with the temporal layer are shown and compared in Table 9. Both are general purpose QA systems using Internet as corpus.

Table 9: Our QA system plus temporal layer compared to START QA system

| | QA + temp layer | | | | START | | | |
|---|---|---|---|---|---|---|---|---|
| | **PREC** | **REC** | **F** | **MRR** | **PREC** | **REC** | **F** | **MRR** |
| **Type 1** | 70.00% | 70.00% | 70.00% | 77.60% | 85.71% | 24.00% | 37.50% | 24.00% |
| **Type 2** | 80.85% | 76.00% | 78.35% | 78.00% | 75.00% | 6.00% | 11.11% | 7.00% |
| **Type 3** | 60.42% | 58.00% | 59.18% | 63.66% | 00.00% | 00.00% | 00.00% | 00.00% |
| **Type 4** | 56.52% | 52.00% | 54.17% | 55.66% | 00.00% | 00.00% | 00.00% | 00.00% |
| **GLOBAL** | 67.02% | 64.00% | **65.47%** | 68.73% | 83.33% | 7.50% | **13.76%** | 7.75% |

The START QA system was only able to answer Types 1 and Type 2 questions. Although, the precision achieved by this system with these types of questions is high, the recall is lower, specially for Type 2 questions (6.00%). Focusing on complex temporal questions (Types 2, 3 and 4), our QA system, which uses temporal information, can be seen to obtain better results than the START QA, which does not use a temporal layer. In conclusion, these results show that the application of a temporal layer improves QA results for complex temporal questions. Concretely, considering the overall results, the QA system using the temporal layer exceeds the START system by a 375.79% as regards the F-measure (48.36% error reduction).

### 6.3 Portability to Other Languages: The Spanish Approach

As said before, the system was initially developed for English but was extended to Spanish as well. Since the task performed by the layer that processes complex questions is language dependent, some adaptation of the system was required: (1) TERSEO for Spanish was used, (2) all the temporal signals stored in the system were translated into Spanish, (3) the question splitter module was adapted to build grammatically correct Spanish *"Cuándo"* ("When") questions.

#### 6.3.1 Decomposition Unit Evaluation for Spanish

The results of the evaluation are shown in Table 10.

Table 10: Evaluation of the system for Spanish

| | **POS** | **ACT** | **CORR** | **PREC** | **REC** | **F** |
|---|---|---|---|---|---|---|
| **TE Identification and Normalization** | 100 | 90 | 82 | 91.1% | 82.0% | 86.3% |
| **Type Identification** | 200 | 200 | 189 | 94.5% | 94.5% | 94.5% |
| **Signal Detection** | 100 | 99 | 97 | 97.9% | 97.0% | 97.4% |
| **Question Splitter** | 100 | 100 | 93 | 93.0% | 93.0% | 93.0% |
| **DECOMPOSITION UNIT** | **200** | **190** | **174** | **91.5%** | **87.0%** | **89.2%** |

Briefly, in the evaluation for Spanish, 174 out of a total of 200 questions were properly processed and decomposed in all aspects (TE identification, type identification, temporal signal detection and splitting, if necessary), which means an F-measure of 89.2% for the whole decomposition process. The best results were obtained by the Signal Detection module (F-measure almost 100%), but the results for Question Splitting and Type Identification (F-measure around 93-94%) and TE Identification and Normalization were also quite good (F-measure around 86%).

The main errors were very similar to the English ones. However, some new problems appeared in Spanish (see Appendix B for details), principally produced by:

- Grammatical errors in the transformation of the second question due to the ambiguity of some words that produces an incorrect POS-tagging. For example: in the expression *"el cometa Hale"* (the Hale comet), the POSTAGGER classifies *"cometa"* (comet) as a verb rather than a noun, which would be the appropriate tag in this case.

- Temporal expressions like *"el año 99"* or *"el 99"* ("year 99"), which in Spanish refer to 1999 in this case, are detected but not resolved. The same problem appears other expressions containing non explicit numeric temporal expressions, i.e. *"el siglo XIX"* ("XIX century"), *"el segundo milenio"* ("second millennium") or the less common word-spelled dates *"mil novecientos noventa y ocho"* ("one thousand nine hundred ninety eight") are not successfully processed by the temporal layer.

- Finally, in questions where the temporal signal is complex, such as: *"un año después de que..."*("a year after..."), both signal detection and question splitting are wrong because this type of complex signal is outside the scope of the system.

#### 6.3.2 Evaluating the influence of the temporal processing in QA systems for Spanish

For this evaluation, the QA system described (Moreda et al., 2008a) was adapted to the Spanish language. As in English, we used the NE-based answer extraction module.

We divided the Spanish evaluation into two experiments, as in the English evaluation:

1. Base QA system evaluation: First the adapted QA system is evaluated without using the temporal layer.

2. Multilayered QA system evaluation: Then the adapted QA system is evaluated when it performs with the temporal layer.

The main aim of this evaluation is to analyze whether the temporal layer can be successfully extended to other languages and deal with other language QA system, like the Spanish-adapted QA system in this case. The 200 temporal question corpus created for the English test were manually translated into Spanish and used in both experiments.

The results obtained by the general purpose Spanish QA system without the temporal layer are shown in Table 11.

The results obtained by the system that has been enhanced with the temporal layer are shown in Table 12.

Table 11: Evaluation of QA system for Spanish temporal questions

| | POS | ACT | CORR | INE | PREC | REC | F | MRR |
|---|---|---|---|---|---|---|---|---|
| **Type 1** | 50 | 35 | 20 | 1 | 57.14% | 40.00% | 47.06% | 45.34% |
| **Type 2** | 50 | 37 | 12 | 0 | 32.43% | 24.00% | 27.59% | 29.06% |
| **Type 3** | 50 | 3 | 0 | 0 | 0.00% | 0.00% | 0.00% | 0.00% |
| **Type 4** | 50 | 4 | 0 | 0 | 0.00% | 0.00% | 0.00% | 0.00% |
| **GLOBAL** | 200 | 79 | 32 | 1 | 40.51% | **16.00%** | 22.94% | 18.60% |

Table 12: Evaluation of QA system plus temporal layer for Spanish temporal questions

| | POS | ACT | CORR | INE | PREC | REC | F | MRR |
|---|---|---|---|---|---|---|---|---|
| **Type 1** | 50 | 35 | 20 | 1 | 57.14% | 40.00% | 47.06% | 45.34% |
| **Type 2** | 50 | 40 | 19 | 0 | 47.50% | 38.00% | 42.22% | 45.96% |
| **Type 3** | 50 | 31 | 15 | 1 | 48.39% | 30.00% | 37.04% | 37.00% |
| **Type 4** | 50 | 31 | 14 | 1 | 45.16% | 28.00% | 34.57% | 34.00% |
| **GLOBAL** | 200 | 137 | 68 | 3 | 49.64% | **34.00%** | 40.36% | 40.58% |

These results for Spanish show, as expected and already proven in the English case, that: a) the QA system enhanced with the temporal layer gives better results in all measures (79.42% improvement in F and 22.60% error reduction in F), and b) the temporal layer is easily extensible to other languages. The final global results are worst than the English approach with this QA system, but this is due to the fact that the baseline results for the Spanish QA system are also worst compared with the English QA system (F-English 38% compared to F-Spanish 23%). In addition, as in the English experiments, an extra experiment with manually corrected Spanish temporal expression identification and normalization is performed and results are shown in Table 13.

Table 13: Evaluation of QA system plus temporal layer for Spanish temporal questions with manually corrected TERN

| | POS | ACT | CORR | INE | PREC | REC | F | MRR |
|---|---|---|---|---|---|---|---|---|
| **Type 1** | 50 | 35 | 20 | 1 | 57.14% | 40.00% | 47.06% | 45.34% |
| **Type 2** | 50 | 43 | 22 | 0 | **51.16%** | **44.00%** | **47.31%** | **51.96%** |
| **Type 3** | 50 | 31 | 15 | 1 | 48.39% | 30.00% | 37.04% | 37.00% |
| **Type 4** | 50 | 31 | 14 | 1 | 45.16% | 28.00% | 34.57% | 34.00% |
| **GLOBAL** | 200 | 140 | 71 | 3 | **50.71%** | **35.50%** | **41.76%** | **42.08%** |

Despite this fact, the Multilayered QA system enhances the performance of the Spanish QA system in all the types of questions, even in the case of complex questions, for which the Base Spanish QA system is unable to find any correct answer. There are very few inexact answers for Spanish as well, thus proving that the system usually obtains exact answers for both languages.

To conclude this second evaluation analysis, extension of the evaluation to Spanish corroborates the conclusions obtained in the English evaluation and also demonstrates that the temporal layer improves the system in the same way regardless of the language.

## 7. Conclusions and Further Work

This study presents a multilayered temporal QA architecture that performs on a multilingual level, in this case English and Spanish. This system processes complex temporal questions by splitting them into simple questions that can be answered by different types of general purpose QA systems. In addition, the system performs a temporal reasoning of the questions with temporal information.

The proposal consists in adding a new layer, on top of a current QA system, which has two main features:

- Complex question decomposition. Questions are decomposed into simple events which generate simple questions (sub-questions) by using the temporal signal that relates the events. The first sub-question (Q-Focus) specifies the type of information the user needs to find. The answer to the second sub-question (Q-Restriction) establishes the temporal restrictions on the list of answers to the Q-Focus. The Q-Focus and the Q-Restriction are the input of a QA system (any type of QA system could be used here).

- Question recomposition. Answers to the Q-Focus and Q-Restriction, obtained from the QA system, are filtered and compared in order to determine their temporal compatibility and construct the final complex answer.

Since the layer that processes complex questions uses lexical and syntactic rules (a grammar), this task is language dependent. Initially, the decomposition unit was prepared for English, but in a very general way. Extension of the system to Spanish was therefore very simple (only some small changes were required), and the same applies to other languages. In addition, the temporal reasoning of the system is performed by TERSEO, which is a multilingual system (now working in Spanish, English, Catalan and Italian) that is easily extensible to any European language.

For evaluation purposes, there were two aims: a) to determine if the decomposition unit processes each type of question properly in order to obtain the appropriate simple factual questions, and b) show how these techniques enhance a general purpose QA system. In order to accomplish these aims, a test bed for English and Spanish was constructed, annotating the question corpus with the correct results for both decomposition and QA tasks, and determining the criteria establishing when a question has been properly decomposed and answered.

The decomposition unit evaluation results for English and Spanish were very good for complex questions (F-measure 89.5% for English and 89.2% for Spanish). When evaluating the performance of the whole multilayered architecture, these results were compared with those obtained by the base QA system without the temporal layer. Great improvement was found, especially in the case of complex questions (Type 3 and 4), in which the base system was not able to answer them at all (4.66% average F-measure for English and 0.00% for Spanish). The multilayered QA system obtained an overall F-measure of approximately 65% for English and 40% for Spanish for all types of questions. Besides, the temporal layer QA system was also compared with an online general purpose QA system called START, demonstrating the difficulty encountered by these general purpose QA systems in answering questions with complex temporal information or temporal relations.

Further work will be done along three main lines of research: 1) resolving the problems detected in the temporal layer after the evaluation process, 2) adding a module to resolve anaphoric co-reference in questions, 3) integrating the event and link information from TIMEML schema (TimeML, 2008) in our system in order to extract a deeper understanding of complex questions, 4) taking into consideration techniques to determine event durations in case of open-ended questions, such as: *"What happened to world oil prices after the Iraqi annexation of Kuwait?"*. For this task, previous work in the field will be considered (Pan et al., 2006b), and 5) applying the layer procedure to other types of complex questions, as well as studying the new features that need to be added to the system to enable it to perform with other languages like Chinese.

## Acknowledgments

This paper has been partially supported by the Spanish government, project TIN-2006-15265-C06-01, and by the framework of the project QALL-ME, which is a 6th Framework Research Programme of the European Union (EU), contract number: FP6-IST-033860.

## Appendix A. Question Decomposition Error Analysis for English

This appendix gives detailed information on the decomposition errors detected in the test for the English language. As shown in Table 5 we distinguish TE identification and normalization, type identification, signal detection and question splitter errors. In Table 14 we specify which questions in the English testbed correspond to which error types. The questions in bold correspond to more than one type of error.

Table 14: Question Decomposition Error Analysis for English

| Error type | testbed question |
|---|---|
| **TE Identification and Normalization** | 81, 83, 89, 92, **97**, **98**, 99, 102, **108**, 112, **114**, 115, **116**, 117, 126, **129**, 133, **135**, **142**, **148** |
| **Type Identification** | **97**, **98**, **108**, **129**, **135**, **148** |
| **Signal Detection** | **101**, **114**, **116**, **129** |
| **Question Splitter** | **101**, 110, **114**, **116**, **129**, **142**, 179, 192 |

The questions implied are listed below. Only erroneous elements are listed and the correct values are indicated in brackets.

<Q id="81"> (ACT: Yes CORR: No)

<QUESTION>Who won the Nobel Peace Prize in '91?</QUESTION>
<TE value="">'91</TE> (CORR value="1991")
</Q>

<Q id="83"> (ACT: Yes CORR: No)
<QUESTION>What tennis player did win Wimbledon women singles in the second millennium year?</QUESTION>
<TE value="">year</TE>(CORR <TE value="2000">second millennium year</TE>)
</Q>

<Q id="89"> (ACT: Yes CORR: No)
<QUESTION>How many planes crashed into Twin Towers in '01?</QUESTION>
<TE value="">'01</TE> (CORR value="2001")
</Q>

<Q id="92"> (ACT: Yes CORR: No)
<QUESTION>What organization was founded in '75 by Bill Gates?</QUESTION>
<TE value="">'75 by</TE> (CORR <TE value="1975">'75</TE>)
</Q>

<Q id="97"> (ACT: No CORR: No)
<QUESTION>What city was the capital of Nicaragua in eighteen fifty-five?</QUESTION>
<TE value=""></TE> (CORR <TE value="1855">eighteen fifty-five</TE>)
<TYPE>1</TYPE> (CORR <TYPE>2</TYPE>)
</Q>

<Q id="98"> (ACT: No CORR: No)
<QUESTION>What was the largest city in Italy in the 17th century?</QUESTION>
<TE value=""></TE> (CORR <TE value="16">the 17th century</TE>)
<TYPE>1</TYPE> (CORR <TYPE>2</TYPE>)
</Q>

<Q id="99"> (ACT: Yes CORR: No)
<QUESTION>Where was Eurovision held in '68?</QUESTION>
<TE value="">'68</TE> (CORR value="1968")
</Q>

<Q id="101"> (ACT: Yes CORR: No)
<QUESTION>Who was the Prime Minister of Spain four years after Jose Maria Aznar presided Spain between 2000 and 2004?</QUESTION>
<SIGNAL>after</SIGNAL> (CORR <SIGNAL>four years after</SIGNAL>)
<Q-FOCUS>Who was the Prime Minister of Spain four years?</Q-FOCUS>
(CORR <Q-FOCUS>Who was the Prime Minister of Spain?</Q-FOCUS>)
</Q>

<Q id="102"> (ACT: No CORR: No)
<QUESTION>Who was the king of Spain after Charles III died in the 1780s?</QUESTION>
<TE value=""></TE> (CORR <TE value="178">the 1780s</TE>)
</Q>

<Q id="108"> (ACT: No CORR: No)
<QUESTION>Who was the president of the US when the AARP was founded five decades ago?</QUESTION>
<TE value=""></TE>(CORR <TE value="195">five decades ago</TE>)
<TYPE>4</TYPE> (CORR <TYPE>3</TYPE>)
</Q>

<Q id="110"> (ACT: Yes CORR: No)
<QUESTION>Who was the Prime Minister of Spain just after the Columbia first flight in the 1980s?</QUESTION>
<Q-FOCUS>Who was the Prime Minister of Spain just?</Q-FOCUS>
(CORR <Q-FOCUS>Who was the Prime Minister of Spain?</Q-FOCUS>)
</Q>

<Q id="112"> (ACT: Yes CORR: No)
<QUESTION>How many members had the European Union when Gladiator was released in '00?</QUESTION>
<TE value="">'00</TE> (CORR <TE value="2000">'00</TE>)

</Q>

<Q id="114"> (ACT: Yes CORR: No)
<QUESTION>What company introduced onto the market a seat with adjustable shoulder support a year before Mariah Carey was born in the 1960s?</QUESTION>
<TE value="">the 1960s</TE> (CORR <TE value="196">the 1960s</TE>)
<SIGNAL>before</SIGNAL> (CORR <SIGNAL>a year before</SIGNAL>)
<Q-FOCUS>What company introduced onto the market a seat with adjustable shoulder support a year?</Q-FOCUS>
(CORR <Q-FOCUS>What company introduced onto the market a seat with adjustable shoulder support?</Q-FOCUS>)
</Q>

<Q id="115"> (ACT: Yes CORR: No)
<QUESTION>Which language was forbidden in Spain during Franco's Dictatorship period 1939-1975?</QUESTION>
<TE value="1975">1939-1975</TE>
(CORR <TE value="1939-1975">1939-1975</TE>)
</Q>

<Q id="116"> (ACT: Yes CORR: No)
<QUESTION>When did Indurain win the Tour a year after the Shawshank Redemption film was released in the 1990s?</QUESTION>
<TE value="">the 1990s</TE> (CORR <TE value="199">the 1990s</TE>)
<SIGNAL>after</SIGNAL> (CORR <SIGNAL>a year after</SIGNAL>)
<Q-FOCUS>When did Indurain win the Tour a year?</Q-FOCUS>
(CORR <Q-FOCUS>When did Indurain win the Tour?</Q-FOCUS>)
</Q>

<Q id="117"> (ACT: Yes CORR: No)
<QUESTION>When did Vesuvius erupt before Sinclair Lewis won Literature Nobel Prize in 1930s?</QUESTION>
<TE value="">1930s</TE> (CORR <TE value="193">1930s</TE>)
</Q>

<Q id="126"> (ACT: Yes CORR: No)
<QUESTION>Who died on a plane crash when Vietnam war was started in late 1960s?</QUESTION>
<TE value="">1960s</TE> (CORR <TE value="1965-1969">late 1960s</TE>)
</Q>

<Q id="129"> (ACT: No CORR: No)
<QUESTION>Who was the king of Spain after Charles IV reigned Spain during the eighteenth century?</QUESTION>
<TE value=""></TE> (CORR <TE value="17">eighteenth century</TE>)
<TYPE>4</TYPE> (CORR <TYPE>3</TYPE>)
<SIGNAL>during</SIGNAL> (CORR <SIGNAL>after</SIGNAL>)
<Q-FOCUS>Who was the king of Spain after Charles IV reigned Spain?</Q-FOCUS>
(CORR <Q-FOCUS>Who was the king of Spain?</Q-FOCUS>)
<Q-REST>When did the eighteenth century happen?</Q-REST>
(CORR <Q-REST>When did Charles IV reign Spain during the eighteenth century?</Q-REST>)
</Q>

<Q id="133"> (ACT: Yes CORR: No)
<QUESTION>What person won the Literature Nobel Prize when James Dean was born in '31?</QUESTION>
<TE value="">'31</TE> (CORR value="1931")
</Q>

<Q id="135"> (ACT: No CORR: No)
<QUESTION>Who was the prime minister of the United Kingdom when the AARP was founded five decades ago?</QUESTION>
<TE value=""></TE> (CORR <TE value="195">five decades ago</TE>)
<TYPE>4</TYPE> (CORR <TYPE>3</TYPE>)
</Q>

<Q id="142"> (ACT: Yes CORR: No)
<QUESTION>Which language was invented by Zamenhof when Berliner patented the Gramophone in the 1880s?</QUESTION>
<TE value="">the 1880s</TE> (CORR <TE value="188">the 1880s</TE>)

<Q-REST>When did Berliner patented the Gramophone in the 1880s happen?</Q-REST>
(CORR <Q-REST>When did Berliner patent the Gramophone in the 1880s?</Q-REST>)
</Q>

<Q id="148"> (ACT: No CORR: No)
<QUESTION>Where was the Woodstock Festival held on August 15 when Unix was developed?</QUESTION>
<TE value=""></TE> (CORR <TE value="XXXX-08-15">August 15</TE>)
<TYPE>4</TYPE> (CORR <TYPE>3</TYPE>)
</Q>

<Q id="179"> (ACT: Yes CORR: No)
<QUESTION>Who was the king of Spain after Charles IV reigned Spain?</QUESTION>
<Q-REST>When did Charles IV reign Spain happen?</Q-REST>
(CORR <Q-REST>When did Charles IV reign Spain?</Q-REST>)
</Q>

<Q id="192"> (ACT: Yes CORR: No)
<QUESTION>Which language was invented by Zamenhof when Berliner patented the Gramophone?</QUESTION>
<Q-REST>When did Berliner patented the Gramophone happen?</Q-REST>
(CORR <Q-REST>When did Berliner patent the Gramophone?</Q-REST>)
</Q>

## Appendix B. Question Decomposition Error Analysis for Spanish

This appendix gives detailed information on the decomposition errors detected in the test for the Spanish language (see table 10)
. In Table 15 we specify which questions correspond to which error types. The questions in bold correspond to more than one type of error.

Table 15: Question decomposition error analysis for Spanish

| Error type | testbed question |
|---|---|
| **TE Identification and Normalization** | **81**, 83, 89, 92, **97**, **98**, 99, **108**, **114**, **116**, **129**, 130, **133**, **135**, 142, 143, 145, 148 |
| **Type Identification** | 2, 6, 9, 31, 45, **81**, **97**, **98**, **108**, **129**, **135** |
| **Signal Detection** | **114**, **116**, **129** |
| **Question Splitter** | 105, 110, **114**, **116**, **129**, **133**, 155 |

The questions implied are listed below. Only erroneous elements are listed and the correct values are indicated in brackets.

<Q id="2"> (ACT: Yes CORR: No)
<QUESTION>¿Durante qué década fue inventado el test del polígrafo?</QUESTION>
<TYPE>2</TYPE> (CORR <TYPE>1</TYPE>)
</Q>

<Q id="6"> (ACT: Yes CORR: No)
<QUESTION>¿En qué año fue lanzado el submarino Nautilus?</QUESTION>
<TYPE>2</TYPE> (CORR <TYPE>1</TYPE>)
</Q>

<Q id="9"> (ACT: Yes CORR: No)

<QUESTION>¿En qué año entró en vigor la enmienda 18?</QUESTION>
<TYPE>2</TYPE> (CORR <TYPE>1</TYPE>)
</Q>

<Q id="31"> (ACT: Yes CORR: No)
<QUESTION>¿Qué año volaron los Wright Brothers por primera vez?</QUESTION>
<TYPE>2</TYPE> (CORR <TYPE>1</TYPE>)
</Q>

<Q id="45"> (ACT: Yes CORR: No)
<QUESTION>¿Qué año fue el gran Incendio de Londres?</QUESTION>
<TYPE>2</TYPE> (CORR <TYPE>1</TYPE>)
</Q>

<Q id="81"> (ACT: No CORR: No)
<QUESTION>¿Quién ganó el Nobel de la Paz en el 91?</QUESTION>
<TE value=""></TE> (CORR <TE value="1991">el 91</TE>)
<TYPE>1</TYPE> (CORR <TYPE>2</TYPE>)
</Q>

<Q id="83"> (ACT: Yes CORR: No)
<QUESTION>¿Qué jugador de tenis ganó Wimbledon mujeres individuales en el año del segundo milenio?</QUESTION>
<TE value="">el año</TE>
(CORR <TE value="2000">en el año del segundo milenio</TE>)
</Q>

<Q id="89"> (ACT: Yes CORR: No)
<QUESTION>¿Cuántos aviones chocaron en las Torres Gemelas en el 01?</QUESTION>
<TE value="">el 01</TE> (CORR value="2001")
</Q>

<Q id="92"> (ACT: Yes CORR: No)
<QUESTION>¿Qué empresa fue fundada en el 75 por Bill Gates?</QUESTION>
<TE value="2008">el 75</TE> (CORR value="1975")
</Q>

<Q id="97"> (ACT: No CORR: No)
<QUESTION>¿Qué ciudad fue la capital de Nicaragua en mil ochocientos cincuenta y cinco?</QUESTION>
<TE value=""></TE> (CORR <TE value="1855">mil ochocientos cincuenta y cinco</TE>)
<TYPE>1</TYPE> (CORR <TYPE>2</TYPE>)
</Q>

<Q id="98"> (ACT: No CORR: No)
<QUESTION>¿Cuál fue la ciudad más grande de Italia en el siglo XVII?</QUESTION>
<TE value=""></TE> (CORR <TE value="16">el siglo XVII</TE>)
<TYPE>1</TYPE> (CORR <TYPE>2</TYPE>)
</Q>

<Q id="99"> (ACT: Yes CORR: No)
<QUESTION>¿Dónde se celebró Eurovisión en el año 68?</QUESTION>
<TE value="2008">el año 68</TE> (CORR value="1968")
</Q>

<Q id="105"> (ACT: Yes CORR: No)
<QUESTION>¿Quién ganó el Nobel de Física cuando el cometa Hale Bopp fue descubierto hace 13 años?</QUESTION>
<Q-REST>¿Cuándo cometió el Hale Bopp fue descubierto hace 13 años?</Q-REST>
(CORR <Q-REST>¿Cuándo fue descubierto el cometa Hale Bopp hace 13 años?</Q-REST>)
</Q>

<Q id="108"> (ACT: No CORR: No)
<QUESTION>¿Quién fue el presidente de los Estados Unidos cuando se fundó AARP hace cinco décadas?</QUESTION>
<TE value=""></TE> (CORR <TE value="195">hace cinco décadas </TE>)
<TYPE>4</TYPE> (CORR <TYPE>3</TYPE>)
</Q>

<Q id=”110”> (ACT: Yes CORR: No)
<QUESTION>¿Quién fue el Presidente de España justo después de que se produjera el primer vuelo del Columbia en los años 80?</QUESTION>
<Q-REST>¿Cuándo se produjo se el primer vuelo del Columbia en los años 80?</Q-REST>
(CORR <Q-REST>¿Cuándo se produjo el primer vuelo del Columbia en los años 80?</Q-REST>)
</Q>

<Q id=”114”> (ACT: No CORR: No)
<QUESTION>¿Qué empresa introdujo en el mercado el primer asiento con respaldo regulable un año antes de que naciera Mariah Carey en los años 60?</QUESTION>
<TE value=””>un año antes</TE> (CORR <TE value=”196”>los años 60</TE>)
<SIGNAL>antes de que</SIGNAL>
(CORR <SIGNAL>un año antes de que</SIGNAL>)
<Q-FOCUS>¿Qué empresa introdujo en el mercado el primer asiento con respaldo regulable un año?</Q-FOCUS>
(CORR <Q-FOCUS>¿Qué empresa introdujo en el mercado el primer asiento con respaldo regulable?</Q-FOCUS>)
</Q>

<Q id=”116”> (ACT: No CORR: No)
<QUESTION>¿Cuándo ganó Indurain el Tour un año después de que se estrenara Cadena Perpetua en los años 90?</QUESTION>
<TE value=””>un año después</TE> (CORR <TE value=”199”>los años 90</TE>)
<SIGNAL>después de que</SIGNAL>
(CORR <SIGNAL>un año después de que</SIGNAL>)
<Q-FOCUS>¿Cuándo ganó Indurain el Tour un año?</Q-FOCUS>
(CORR <Q-FOCUS>¿Cuándo ganó Indurain el Tour?</Q-FOCUS>)
</Q>

<Q id=”129”> (ACT: No CORR: No)
<QUESTION>¿Quién fue el Rey de España después de que Carlos IV reinara España durante el siglo XVIII?</QUESTION>
<TE value=””></TE> (CORR <TE value=”17”>el siglo XVIII</TE>)
<TYPE>4</TYPE> (CORR <TYPE>3</TYPE>)
<SIGNAL>durante</SIGNAL> (CORR <SIGNAL>después</SIGNAL>)
<Q-REST>¿Cuándo fue el siglo XVIII?</Q-REST>
(CORR <Q-REST>¿Cuándo reinó Carlos IV España durante el siglo XVIII?</Q-REST>)
</Q>

<Q id=”130”> (ACT: Yes CORR: No)
<QUESTION>¿Quién ganó Wimbledon femenino individuales antes de que Rafa Nadal ganara Wimbledon este año?</QUESTION>
<TE value=””>este año</TE> (CORR <TE value=”2008”>este año</TE>)
</Q>

<Q id=”133”> (ACT: Yes CORR: No)
<QUESTION>¿Qué persona ganó el premio Nobel de Literatura cuando James Dean nació en el año 31?</QUESTION>
<TE value=””>el año 31</TE> (CORR value=”1931”)
<Q-REST>¿Cuándo jamó Dean nació en el año 31?</Q-REST>
(CORR <Q-REST>¿Cuándo nació James Dean en el año 31?</Q-REST>)
</Q>

<Q id=”135”> (ACT: No CORR: No)
<QUESTION>¿Quién fue el Presidente de Reino Unido cuando AARP fue fundada hace cinco décadas?</QUESTION>
<TE value=””></TE> (CORR <TE value=”195”>hace cinco décadas</TE>)
<TYPE>3</TYPE> (CORR <TYPE>4</TYPE>)
</Q>

<Q id=”142”> (ACT: Yes CORR: No)
<QUESTION>¿Qué lengua fue inventada por Zamenhof cuando Berliner patentó el disco de vinilo en la década de 1880?</QUESTION>
<TE value=”1880”>1880</TE>
(CORR <TE value=”188”>la década de 1880</TE>)
</Q>

<Q id="143"> (ACT: No CORR: No)
<QUESTION>¿Dónde se celebrarán las Olimpiadas cuando Polonia adopte el Euro en la década de 2010?</QUESTION>
<TE value=""></TE> (CORR <TE value="201">la década de 2010</TE>)
</Q>

<Q id="145"> (ACT: Yes CORR: No)
<QUESTION>¿Cuándo ganó Gary Becker el premio Nobel de Economía antes de que Zapatero fuera elegido Presidente de España en los últimos años?</QUESTION>
<TE value="">los últimos años</TE> (CORR value="[2003-2008]")
</Q>

<Q id="148"> (ACT: No CORR: No)
<QUESTION>¿Dónde se celebró el Festival de Woodstock el 15 de agosto cuando el Unix fue desarrollado?</QUESTION>
<TE value=""></TE> (CORR <TE value="XXXX-08-15">el 15 de agosto</TE>)
</Q>

<Q id="155"> (ACT: Yes CORR: No)
<QUESTION>¿Quién ganó el Nobel de Física cuando el cometa Hale Bopp fue descubierto?</QUESTION>
<Q-REST>¿Cuándo cometió el Hale Bopp fue descubierto?</Q-REST>
(CORR <Q-REST>¿Cuándo fue descubierto el cometa Hale Bopp?</Q-REST>)
</Q>